\documentclass[utf8]{FrontiersinHarvard} 

\usepackage{url,hyperref,microtype,subcaption}
\usepackage[onehalfspacing]{setspace}

\def\keyFont{\fontsize{8}{11}\helveticabold }
\def\firstAuthorLast{Liu and Wang {et~al.}} 
\def\Authors{Bangyuan Liu\,$^{1,\dagger}$, Tianyu Wang\,$^{2,\dagger}$, Velin Kojouharov\,$^{2}$, Frank L. Hammond III\,$^{1,*}$, and Daniel I. Goldman\,$^{2,*}$}
\def\Address{$^{1}$ARMLab, School of Mechanical Engineering, Georgia Institute of Technology, Atlanta, Georgia, USA \\
$^{2}$CRAB Lab, School of Physics, Georgia Institute of Technology, Atlanta, Georgia, USA \\
$^{\dagger}$These two authors contributed equally to this work}
\def\corrAuthor{Frank L. Hammond III}

\def\corrEmail{frank.hammond@me.gatech.edu}

\begin{document}
\onecolumn
\firstpage{1}

\title {Robust self-propulsion in sand using simply controlled vibrating cubes} 

\author[\firstAuthorLast ]{\Authors} 
\address{} 
\correspondance{} 

\extraAuth{Daniel I. Goldman \\ daniel.goldman@physics.gatech.edu}

\maketitle

\begin{abstract}
Much of the Earth and many surfaces of extraterrestrial bodies are composed of in-cohesive particle matter. Locomoting on granular terrain is challenging for common robotic devices, either wheeled or legged. In this work, we discover a robust alternative locomotion mechanism on granular media -- generating movement via self-vibration. To demonstrate the effectiveness of this locomotion mechanism, we develop a cube-shaped robot with an embedded vibratory motor and conduct systematic experiments on diverse granular terrains of various particle properties. We investigate how locomotion changes as a function of vibration frequency/intensity on granular terrains. Compared to hard surfaces, we find such a vibratory locomotion mechanism enables the robot to move faster, and more stable on granular surfaces, facilitated by the interaction between the body and surrounding granules. The simplicity in structural design and controls of this robotic system indicates that vibratory locomotion can be a valuable alternative way to produce robust locomotion on granular terrains. We further demonstrate that such cube-shape robots can be used as modular units for morphologically structured vibratory robots with capabilities of maneuverable forward and turning motions, showing potential practical scenarios for robotic systems.

\tiny
 \keyFont{ \section{Keywords:} Vibration, granular media, robot, locomotion, modular robot, robophysics} 
\end{abstract}

\section{Introduction}

Many terrestrial and extraterrestrial landmasses are composed of soft flowable material, like sand and snow. Among them, granular media, a collection of discrete, solid particles that exhibit energy dissipation upon interaction, is challenging for conventional wheeled and tracked devices to locomote on. To surmount this challenge, diverse robotic systems have been developed to produce controllable movement in granular materials. These systems encompass legged configurations \citep{liang2012amphihex,qian2015principles,zhong2018cpg,lee2022compact}, limbless structures \citep{maladen2011undulatory,marvi2014sidewinding}, and rover designs \citep{knuth2012discrete,shrivastava2020material}. 
A common issue faced by diverse robotic systems attempting locomotion within granular media lies in the potential transition of the granular terrain from a solid to a flowing state, which occurs when the force per unit area exerted by the robot exceeds its yield stress or input energy threshold. The intricate interaction between the robot body and the terrain often engenders a coupled relationship encompassing the robot's movement, the resistive force the robot experiences, and the dynamic changing of the terrain state before, during, and after the robot's movement. Particularly in cases involving shear-based locomotion, the granular material can aggregate into formations that act as obstacles, hindering the robot's progress.


A vibration-driven robot leverages periodic oscillations or rotations of internal components to generate motion throughout the entire body~\citep{golitsyna2018periodic,calisti2017fundamentals,reis2013morphological,becker2011modeling}. In contrast to conventional wheeled, legged, or elongated limbless robots, vibration-driven robots offer a simpler and more compact body design, without the need for body deformation and exposed actuators or joints. This intrinsic feature effectively prevents potential damage or malfunctions to the robot arising from particle infiltration or obstruction of the actuation mechanisms. Further, such mechanism and design allow for a greater contact area between the robot's body and the terrain. Therefore, the yield strain and stress for supporting body weight and producing motion propulsion can be reduced; consequently, the risk of terrain collapse is mitigated.

Previous research has designed and studied vibration-driven robots that can locomote on various terrains, such as hard surfaces~\citep{li2021programming,notomista2019study,rubenstein2012kilobot}, fluid surfaces~\citep{cocuzza2021vibration}, and amphibious environments~\citep{wang2022miniature, tang2018speedy}. These vibratory robots typically utilize a consistent actuation principle, wherein vibratory actuators induce multi-directional movement through alterations in force orientation, and the accumulated movement direction is dictated by anisotropic friction or resistive forces.
These forces stem from specific directional components of periodic internal forces or inherent body features~\citep{golitsyna2018periodic,calisti2017fundamentals}. However, the performance of such a vibration-driven mechanism on granular media remains unexplored.

\begin{figure}[t]
\begin{center}
\includegraphics[width=0.7\textwidth]{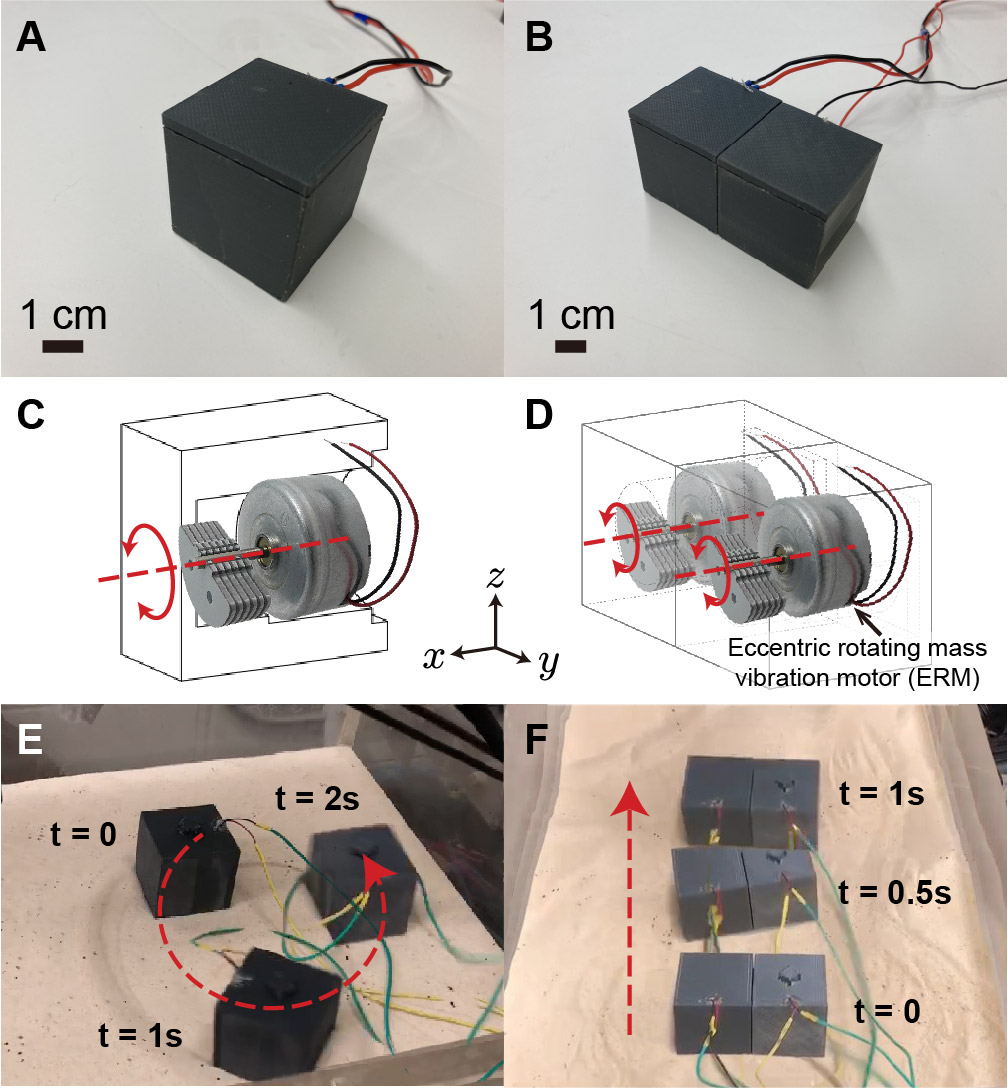}
\end{center}
\caption{Vibratory single cube and bi-cube fabrication and motion. Figure (A), (C), and (E) shows the appearance, perspective drawing, and the turning maneuver of a single cube. By combining two single cubes together, the bi-cube is fabricated. Figure (B), (D), and (F) shows the appearance, perspective drawing, and the forward-moving maneuver of a bi-cube. The locomotion of single cube and bi-cube are recorded in supplementary video S1.}\label{fig:1}
\end{figure}

Due to its design simplicity of structure, ease of control, and enlarged body-terrain contact area, a vibratory presents a promising avenue for achieving robust and effective locomotion on granular media. This work introduces the development of a novel vibrating robotic system (Fig. \ref{fig:1}) to investigate the locomotion capabilities of such vibration mechanisms within different granular mediums and identify potential practical applications.

\section{Materials and Methods}

\subsection{Robot Design and Maneuver manipulation}
\subsubsection{Single Cube}
For design simplicity, we used Solidworks to create a cubic box with an open side for the motor and a press-fit lid to close the cube (as shown in Fig. \ref{fig:1}). The 4-cm-side-length cube was then 3D printed using a LulzBot TAZ Workhorse 3D Printer and polylactic acid (PLA) as the printing material. 
To generate the vibration, we selected an eccentric rotating mass vibration motor (ERM), specifically the VJQ24 from Vybronics Inc (weight 31 g, 2550 rpm at 5 V DC). The rotary axis of the vibration motor lies horizontally in the x direction, parallel to the cube's bottom surface. When power is applied, the uneven mass starts rotating, which leads to the rotary oscillation around x axis. By switching the voltage from positive to negative, the vibration motor rotation direction converts from counter-clockwise to clockwise (viewing in the positive x direction), which allows the single cube and bi-cube to generate maneuvers. The position of the vibration motor is adjusted to the cube's center, guaranteeing alignment between the center of the cube's mass and its geometric center. Inside the cube, we implemented a structure to securely press fit the motor in place to ensure that it remains stationary while vibrating. 

\begin{figure}[t]
\begin{center}
\includegraphics[width=0.8\textwidth]{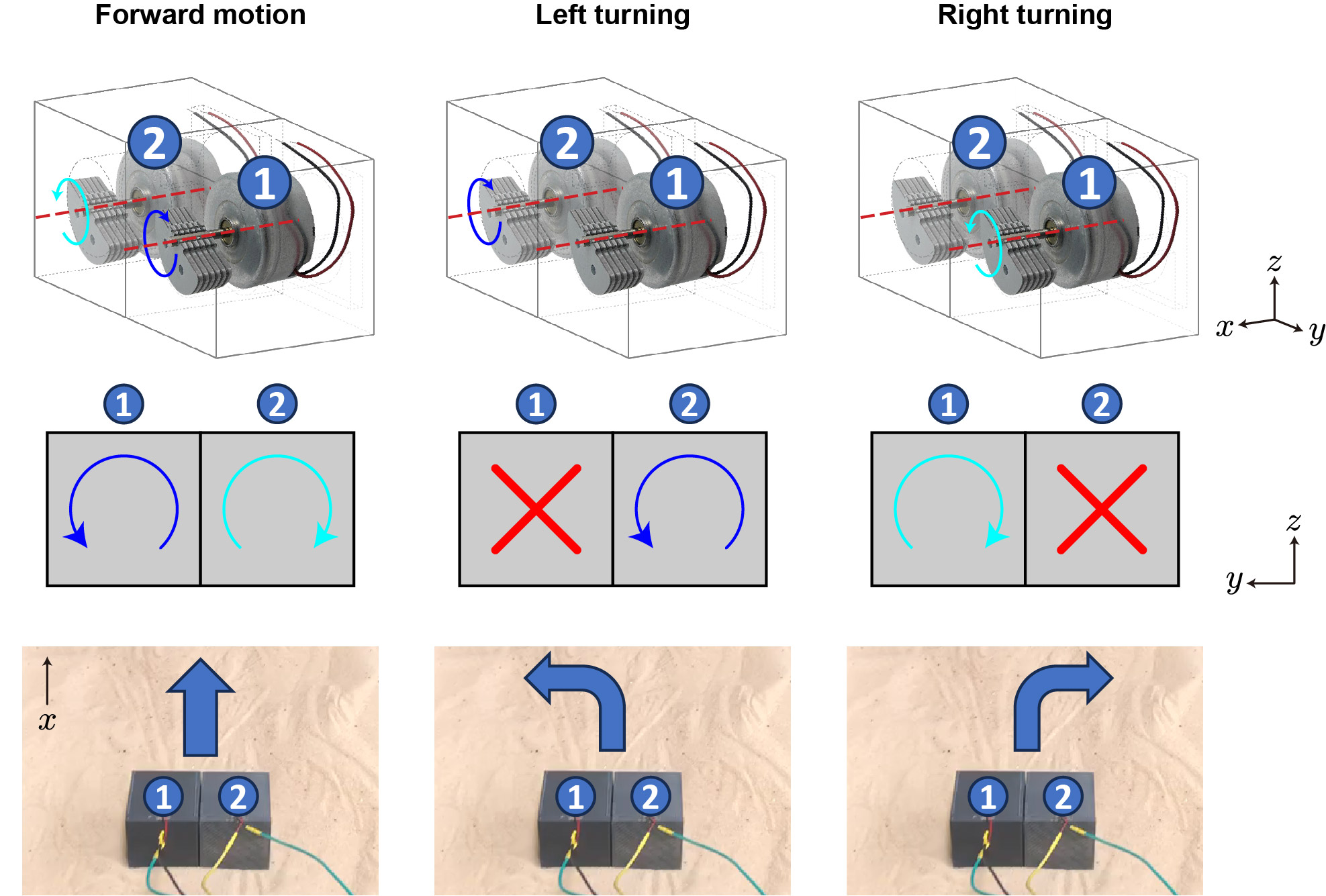}
\end{center}
\caption{Bi-cube steering mechanism. The bi-cube can execute three maneuvers by controlling the rotational state of vibration motors \#1 and \#2 inside the left and right cubes: forward motion, left turning, and right turning, from the left column to the right column, with each row illustrating the motor state from the perspective view, back view, and the corresponding maneuver.}
\label{fig:2}
\end{figure} 

Influenced by the simultaneous lateral oscillatory inertia generated by the single cube system, the cube forward locomotion is accompanied by turning (shown in Fig. \ref{fig:1}). When the input voltage is positive, the counter-clockwise-rotating vibration motor induces a leftward turning, while negative input voltage induces a rightward turning.

\subsubsection{Bi-cube and maneuver control}
The bi-cube robot is formed by two firmly bonded identical cubes by combining two cubes in the orientation in which two ERM's rotary axes are parallel to each other. Such a bi-cube robot can not only move forward straight but can also turn. 

Based on the maneuver test, when the left cube (marked as \#1 in Fig.~\ref{fig:2}) vibration motor rotates counter-clockwise and the right cube (marked as \#2) rotates clockwise, the lateral oscillation influence of each motor can be canceled out, thus the bi-cube robot performs a forward motion. When \#1 cube turns off and \#2 cube rotates counter-clockwise, the bi-cube robot performs a left turn. Similarly, when \#2 cube turns off and \#1 cube rotates clockwise, the bi-cube robot performs a right turn. For the other manipulation combination case, we did not observe stable maneuvers occur. 

Additionally, the forward maneuver requires that the input voltage magnitude applied to both Cube \#1 and Cube \#2 be the same, which helps the cubes to synchronize and resonate. A small divergence between two sides' voltage would influence the maneuver performance greatly, and finally, lead to unstable arbitrary motion.

\begin{figure}[t]
\begin{center}
\includegraphics[width=0.7\textwidth]{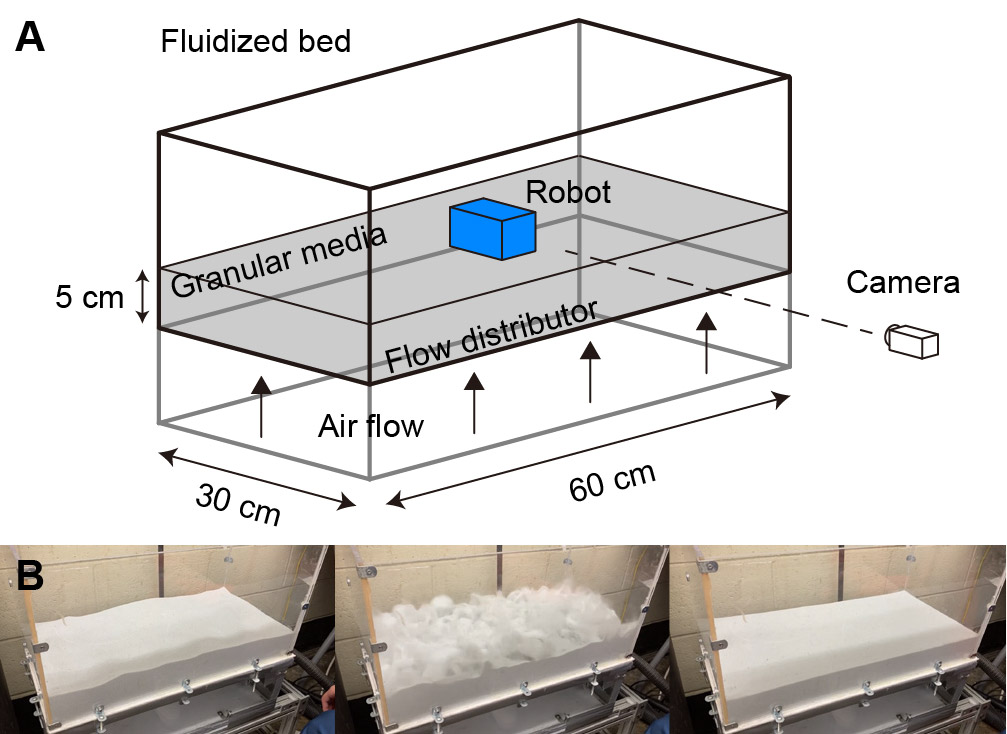}
\end{center}
\caption{The automated terrain creation and locomotion testing system. (A) Diagram of the air-fluidized bed for the robot locomotion testing. (B) the process of flattening the granular media surface before each experiment and creating a loosely packed state.}\label{fig:3}
\end{figure}

\subsection{Experiment Environment}
\subsubsection{Air-fluidized Testbed}
To ensure consistent initial conditions for our experiments on granular media, we implemented a terrain creation and locomotion testing system (as shown in Fig. \ref{fig:2}), following the approach described in our previous work \citep{qian2013automated}. The system uses an air-fluidized bed, a container (60cm long, 30cm wide) filled with granular material (5cm deep) such that the surface can be flattened at the beginning of each experiment by blowing air through a porous rigid plastic layer. A vacuum (Vacmaster) is used to blow the bed. The flow distribution layer, which evenly distributes the air across the bed, is approximately 0.5 cm thick and has randomly distributed pores with a diameter of 50 micrometers. This allows for precise control over the bed's surface and ensures that the initial conditions of each experiment are consistent.

\subsubsection{Granular materials}
We expect that the locomotor capabilities of our robotic cube can vary depending on the type of granular material it is tested on, as the particle sizes of different materials can affect its movement. To investigate this, we tested robot speed and energy use on three different types of granular material in our experiments: glass particles with an average diameter of approximately 200 micrometers, sand with particle sizes ranging from 500 to 700 micrometers (fine granular media), and 1000 to 1200 micrometers (coarse sand, collected in Yuma County, Arizona, USA). As a comparison, we also tested on a hard wooden board (as shown in Fig. \ref{fig:3}D).

\subsection{Experiment Protocol}
For each of the four terrains we considered, we carried out a series of experiments with the bi-cube robot. We varied the input voltage to the system from 0 to 8V to test its forward speed and cost of transport (COT). Specifically, we limited the robot's rotational movement using two parallel walls that were 10 cm away from each other (slightly larger than the robot's width 8 cm). For each voltage input value, we carried out three repeated experiments. In each experiment, we first flattened the granular material surface, then placed the robot at one end of the tank ($>$5 cm away from the wall) and recorded the robot's locomotion from the robot started vibrating to the robot reached the other end of the tank. For the experiments where the robot was unable to reach the other end of the tank, we kept the robot running for at least 30 seconds. To capture the robot's locomotion trajectory and poses, we implemented a motion tracking system (OptiTrack) using 4 OptiTrack Flex 13 cameras. We attached infrared reflective markers to the robot body and tracked their 3D positions using the system at a 120 fps frame rate, thus the progress of the locomotion can be fully reconstructed and analyzed. We tracked the rigid body position and orientation in the world frame to calculate the forward speed. To calculate the cost of transport, we measured the power consumption of the system using an INA260 precision digital power sensor that monitors power at a 100 Hz frequency.

\section{Result and Discussion}
\subsection{Locomotor Performance}
\subsubsection{Forward velocity}

\begin{figure}[t]
\begin{center}
\includegraphics[width=1\textwidth]{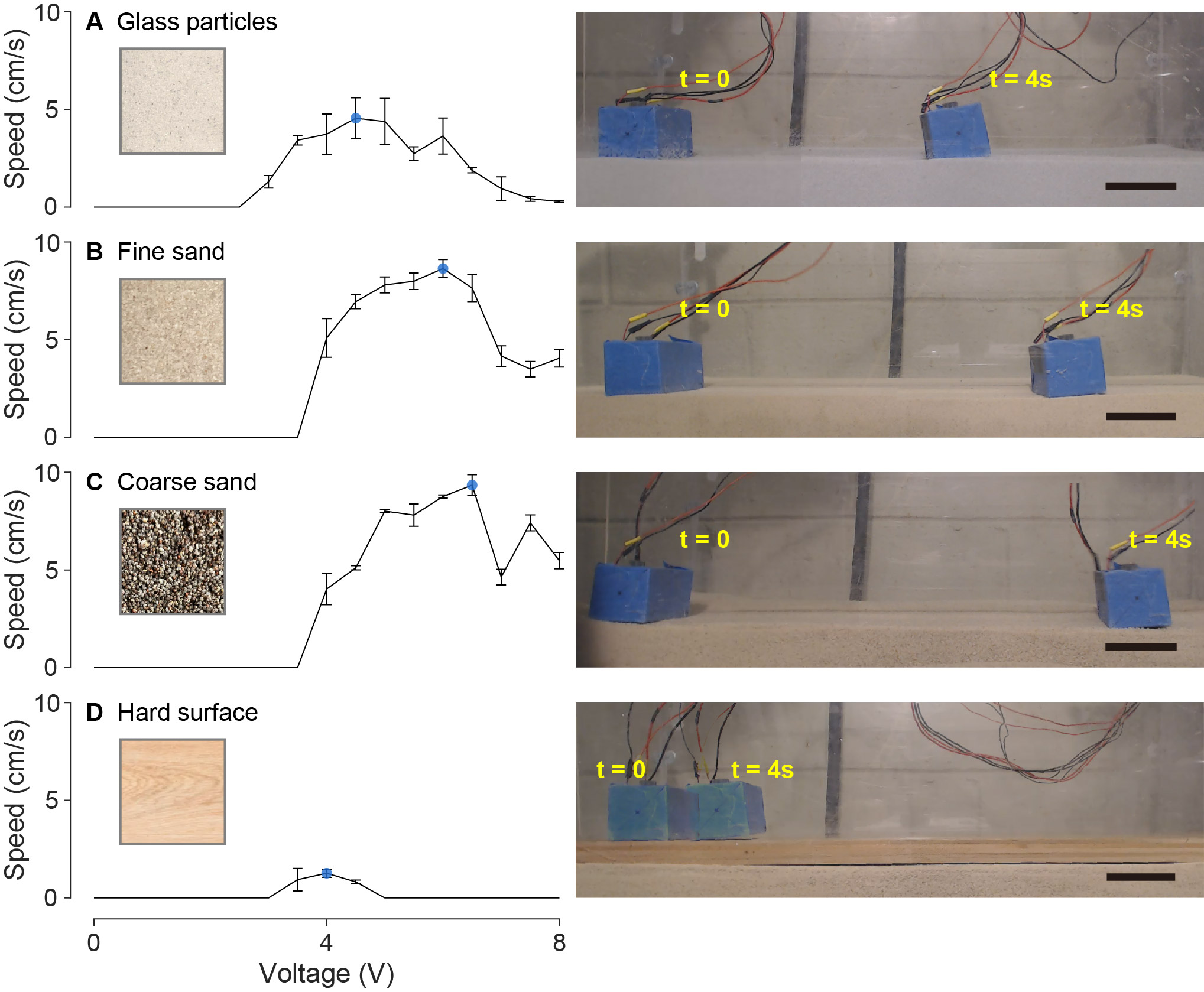}
\end{center}
\caption{Bi-cube velocity test on (A) glass particles, (B) fine granular media (fine sand), (C) coarse granular media (coarse sand), and (D) hard wooden surface. Each figure in the left column shows the averaged velocity of three experiment trials as a function of input voltage from 0 to 8V, with the error bar showing the standard derivation. The local maximum velocity on each terrain is marked by the blue dot. The right column shows an example locomotion of the robot operated with the optimal input voltage in each terrain, in which the frames are recorded at t=0 and t=4s. The black scale bars indicate 5cm. The performance are recorded in supplementary video S2.}\label{fig:4}
\end{figure}

We measured the average speed under various applied voltages to quantify the bi-cube robot's locomotion performance. Utilizing tracked data, we computed the average forward speed (measured in cm/s) achieved by the bi-cube robot during each experiment. The values were averaged over three separate trials, as depicted in Fig.~\ref{fig:4}, where error bars represent the standard deviation.

In glass particles, the bi-cube robot cannot generate effective locomotion when the input voltage is less than 3V. With increasing input voltage, the forward locomotion speed increases, reaching the local maximum of $4.55 \pm 1.05$ cm/s at 4.5V. As the input voltage steadily increases, reaching the maximum voltage of the ERM at 8V, performance gradually declines. Comparable trends in locomotion performance are observed in fine granular media (fine sand) and coarse granular media (coarse sand) as well. However, the minimum input voltage required for effective motion increases to 4.5V in these cases. Moreover, local maxima in forward speed emerge at higher input voltages. Specifically, in fine granular media (fine sand), the local maximum is observed at $8.64 \pm 0.46$ cm/s with an input voltage of 6V. In the case of coarse granular media (coarse sand), the local maximum reaches $9.34 \pm 0.54$ cm/s with an input voltage of 6.5V. For comparative analysis, we conducted locomotion performance tests on a hard wooden surface. The results illustrate that only a narrow input voltage range (from 3.5V to 4.5V) yields forward movement for the bi-cube robot on this terrain. Further, the maximum speed achieved is only $1.27 \pm 0.21$ cm/s at 4V, significantly slower than the velocities achieved in granular media.

Through these experimental findings, we validate that the bi-cube robot we developed is capable of achieving effective locomotion across various granular media types (at rates greater than 1 body length per second). Further, it is observed that the optimal operating voltage increases as the grain size increases.

\subsubsection{Locomotion efficiency}
\begin{figure}[t]
\begin{center}
\includegraphics[width=0.75\textwidth]{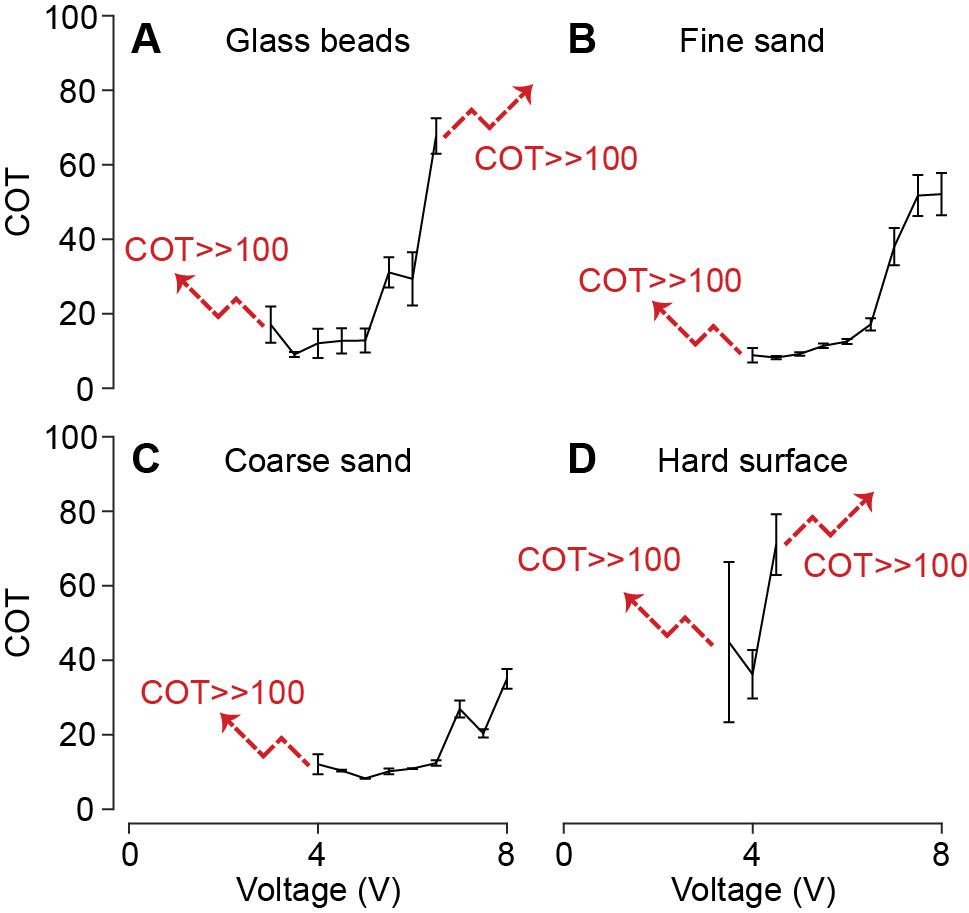}
\end{center}
\caption{Bi-cube locomotion efficiency test on (A) glass particles, (B) fine granular media (fine sand), (C) coarse granular media (coarse sand), and (D) a hard wooden surface.  Each figure shows the averaged cost of transport (COT) of three trials as a function of input voltage from 0 to 8V, with the error bar showing the standard derivation. The red zigzag dash arrow marks the region where the cube does not move effectively, and thus COT rises tremendously.}\label{fig:5}
\end{figure}

In addition to evaluating the forward locomotion speed, we conduct real-time power consumption measurements using an INA260 power sensor for each experiment. Subsequently, we calculate the cost of transport (COT) following $\text{COT} = \frac{P}{mgv}$, where $P$ and $v$ represent the average power consumption and speed achieved during the locomotion process, respectively, and $m$ is the mass of the bi-cube robot (147 g). Fig.~\ref{fig:5} shows COT values of the bi-cube robot traversing four different types of terrain that we tested. 

Experiment results reveal that the local minima of COT vary corresponding to grain sizes. Notably, this minimal cost of transport values remains below 10 across all granular terrains: $9.06 \pm 0.66$ at 3.5V in glass particles, $8.23 \pm 0.42$ at 4.5V in fine granular media (fine sand), and $8.31 \pm 0.09$ at 5V in coarse granular media (coarse sand). However, we notice that the local minima of the cost of transport do not coincide with the local maxima of speed. Note that we have excluded data points corresponding to input voltages where the robot is unable to achieve effective locomotion, resulting in a cost of transport exceeding 100. Our findings demonstrate that the bi-cube robot demonstrates more efficient locomotion within granular media in comparison to hard surfaces. This highlights its substantial potential for broader applications within granular environments.

\subsection{Motion recording and analysis}
To gain insight into the mechanisms governing the bi-cube robot's translation through vibration and subsequently develop either a kinematic or dynamic model to elucidate its motion, we utilized a high-speed 240 fps camera to capture the rigid body orientations of the robot. Subsequently, we extracted specific frames to calculate the instantaneous displacements in both the forward ($x$) and gravity ($z$) directions, as well as the rotational motion within the xz plane ($\theta$).

\begin{figure}[t]
\begin{center}
\includegraphics[width=1\textwidth]{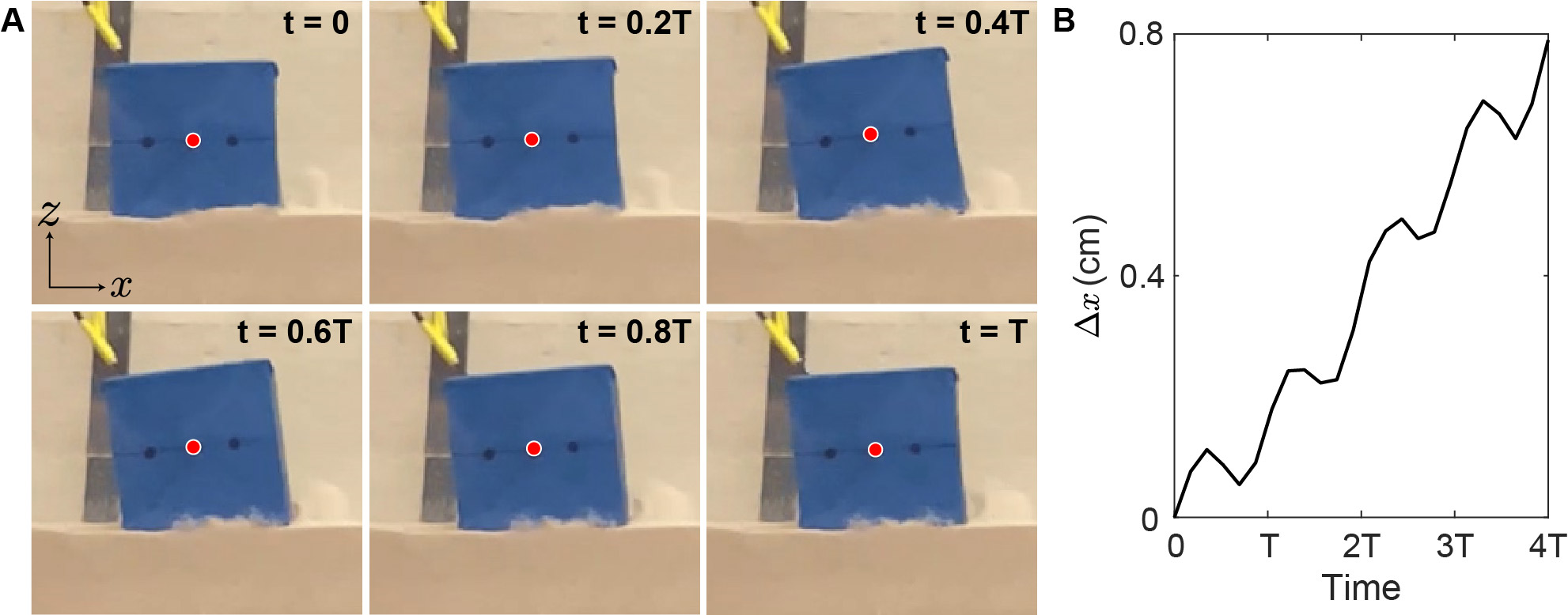}
\end{center}
\caption{Bi-cube motion tracking. (A) A sequence of the robot posture in the abstracted 2D workspace over one complete period, the red dot represents the center of geometry. Slow motion record is provided in supplementary video S3. (B) The periodic trajectories of $x$ displacement of the center of geometry.}\label{fig:6}
\end{figure}

In Fig.~\ref{fig:6}A, we present a sequence of the robot's postures in a period of motion in the quasi-2D setup under 5V input. Fig.~\ref{fig:6}B shows the $x$ displacements of the center of geometry. The $x$ trajectory exhibits periodic behavior with a pattern: over the course of a single motion cycle, the robot experienced two distinct phases. In one half of the cycle, the robot moved backward, while in the other half, it moved forward. A net forward displacement was achieved because the distance covered during the forward movement phase is greater than that during the backward movement phase. 

\subsection{Slope climbing}
Slope climbing remains challenging for many robots designed to navigate granular environments~\citep{shrivastava2020material}, as steep granular slopes tend to be sensitive to stress and shear forces disturbances \citep{gravish2014effect,tegzes2003development}, which can cause avalanche. The intricate nature of these terrains becomes particularly evident during climbing maneuvers, as even minor disturbances introduced by robot motion can trigger avalanches and instigate the transition of the terrain from a solid to a fluid-like state, resulting in failure.

\begin{figure}[t]
\begin{center}
\includegraphics[width=1\textwidth]{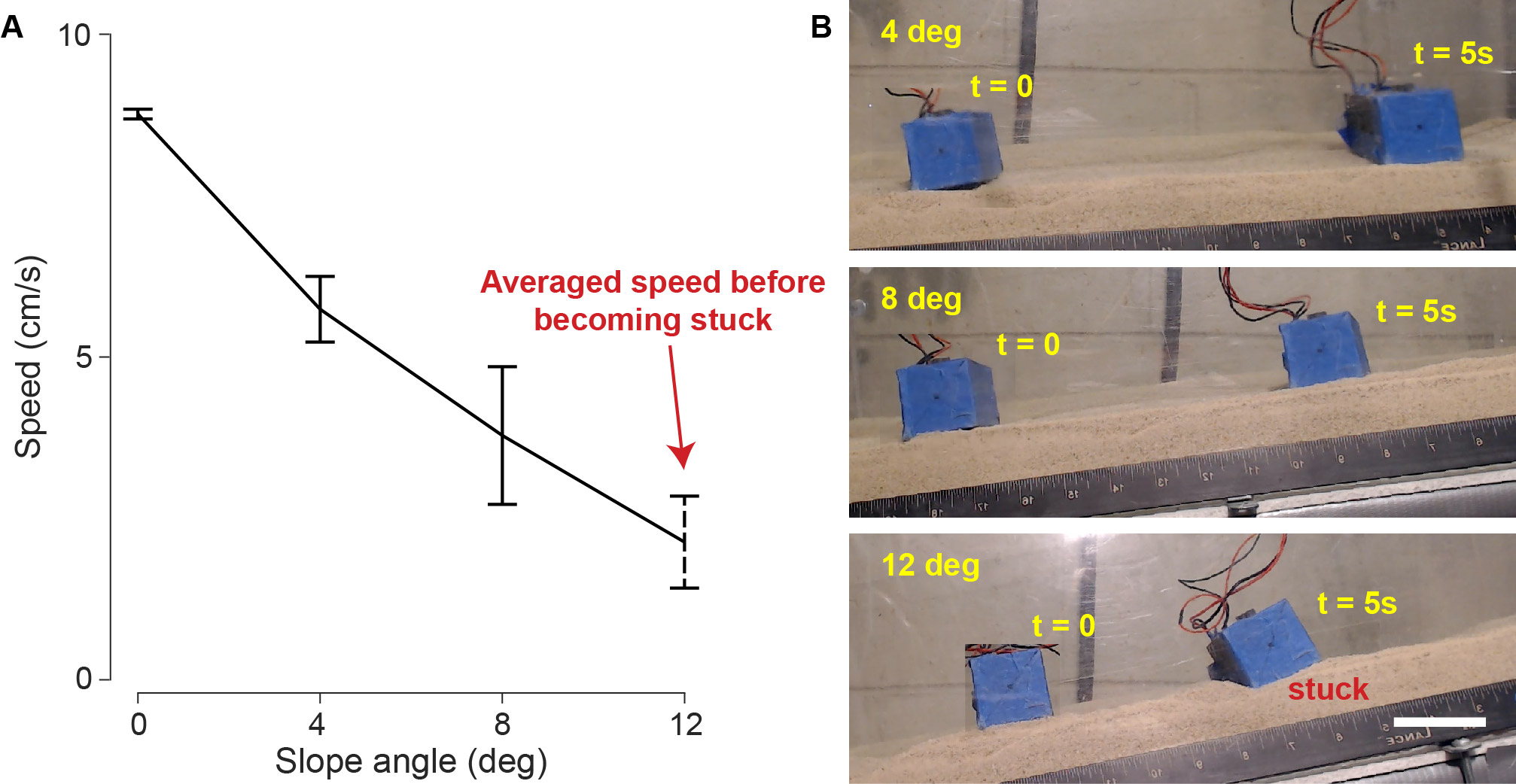}
\end{center}
\caption{Bi-cube slope climbing. (A) The averaged velocity of three trials versus slope angle with standard derivation bar. The cube is tested on coarse granular media (coarse sand) at 6V. At 12 12-degree slope, we recorded the average speed before the bi-cube got stuck. (B) The bi-cube slope climbing motion (4, 8, 12 degrees), recorded at frames of $t=0$ s and $t=$5 s. The scale bar is 5 cm. The performance is recorded in supplementary video S4.}\label{fig:7}
\end{figure}

To investigate the slope-climbing capabilities of the bi-cube robot, we conducted tests on the testbed with coarse granular media (coarse sand, 1000 to 1200 micrometers), with one side of the testbed elevated, thereby forming granular slopes with 4, 8, and 12 degrees of inclined angle. Fig.~\ref{fig:7}A shows the average speed and standard deviation of the cube's motion, powered by 6V input voltage, averaged over three trials. With increasing slope angles, there is a continuous decrement in the cube's velocity. At a 12-degree slope angle, we observed that the relatively gentle slope would sometimes lead to collapse during robot locomotion. Consequently, the robot would become stuck shortly after climbing a few body lengths and be unable to escape.

\subsection{Escaping from sticking}
Given that the testbed is fluidized before each experiment, the loosely compacted granular terrain can unpredictably collapse during robot locomotion. This can lead to the formation of pits in the terrain which can cause the robot to get stuck. Thus we carried out experiments to test the bi-cube robot's ability to escape from pits. Through experiments, we verified that with appropriate input voltages (e.g., 5V in fine granular media), the bi-cube robot can manage to extricate itself from a pit: The robot first engaged in a process of crawling, gradually moving the sand pile from its front to rear; and eventually, this enabled the robot to successfully exit. We provide a demonstration in supplementary video S5, in which the robot first got stuck in a pit in fine granular media (fine sand) when actuated by 3V, and escaped with a 5V input. 

\subsection{Maneuverability test}
\begin{figure}[t]
\begin{center}
\includegraphics[width=0.7\textwidth]{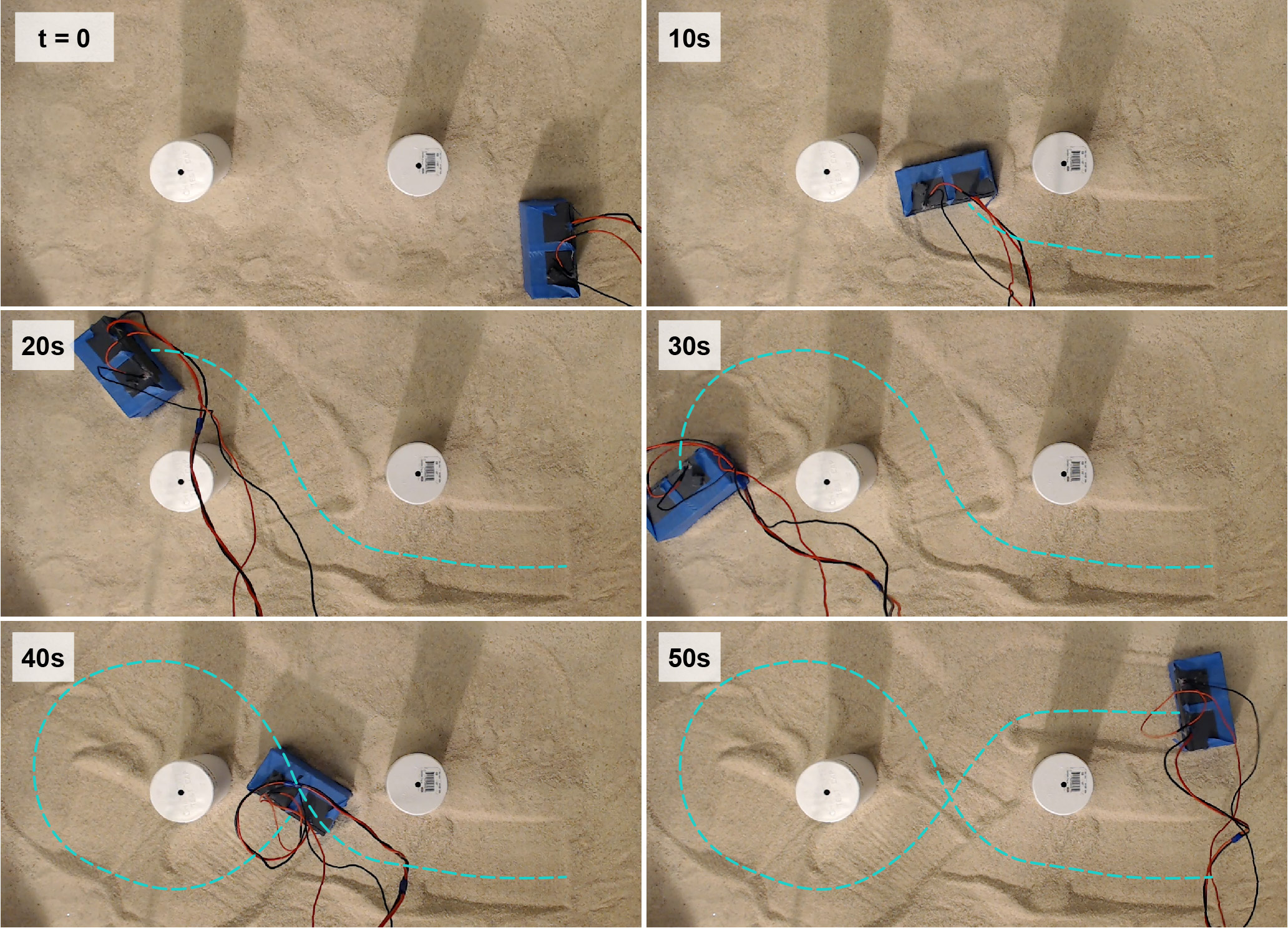}
\end{center}
\caption{Bi-cube maneuver demonstration. The cube moves in an '$\alpha$'-shaped trajectory (marked by the cyan dash line) around cylinder obstacles. The cube maneuver is manually controlled (switched between maneuvers as shown in Fig.~\ref{fig:2}). Each frame records the cube position at $t=0, 10, 20, 30, 40, 50$ s.}\label{fig:8}
\end{figure}

The bi-cube robot exhibits the capacity to execute forward, left, and right turning maneuvers, enabling effective navigation across 2D granular terrains. We illustrate this capability through a maneuver demonstration conducted on a coarse granular media surface (comprising particles ranging from 1000 to 1200 micrometers). Notably, two cylindrical obstacles have been rigidly placed within the terrain, as depicted in Fig.~\ref{fig:8}. In this demonstration, the robot's maneuvering actions are manually switched among forward moving, left turning and right turning as shown in Fig.~\ref{fig:2}. The cube follows an `$\alpha$'-shaped trajectory, avoiding any potential collisions with the cylinder obstacles. We provide supplementary video S6, which describes the bi-cube robot's agility and maneuverability while traversing complex granular terrains.

\section{Conclusion}
In this paper, we systematically tested the capability of vibratory locomotion on the granular medium based on experiments. The vibration cube exhibits the capability to navigate across granular terrains of various particle sizes as well as solid ground. Strikingly, compared to the hard ground, such vibratory locomotion method outperforms on granular media surfaces both in velocity and efficiency. We posit that the flowable nature of the granular terrain plays a pivotal role in stabilizing motion and attenuating extraneous vibration energy, thereby amplifying locomotive efficacy. The inherent vibratory locomotion mechanism showcases an inherent affinity for granular terrains, suggesting a harmonious alignment between the mechanism and such environments.

A solitary cube only demonstrates the capacity to execute left and right turns. However, through the fusion of two individual cubes, a bi-cube configuration is achieved, facilitating forward, left, and right turning. This amalgamation imparts a notable enhancement to maneuverability. Moreover, the inherent simplicity of the vibration cube underscores its potential to exhibit swarming capabilities on granular terrain.

Future work includes revealing the granular vibratory locomotion mechanism via both comprehensive theoretical modeling and experimental validation. Besides, we intend to conduct an in-depth investigation into the influence of particle size and density on locomotion efficiency based on simulation. Additionally, we will upgrade the vibration cube into a swarm robotic system, integrating self-feedback loops and inter-unit communication, for potential future application in exploration.

\bibliographystyle{Frontiers-Harvard} 
\bibliography{main}

\begin{thebibliography}{21}
\providecommand{\natexlab}[1]{#1}
\expandafter\ifx\csname urlstyle\endcsname\relax
  \providecommand{\doi}[1]{doi:\discretionary{}{}{}#1}\else
  \providecommand{\doi}{doi:\discretionary{}{}{}\begingroup \urlstyle{rm}\Url}\fi
\providecommand{\selectlanguage}[1]{\relax}
\providecommand{\bibAnnoteFile}[1]{%
  \IfFileExists{#1}{\begin{quotation}\noindent\textsc{Key:} #1\\
  \textsc{Annotation:}\ \input{#1}\end{quotation}}{}}
\providecommand{\bibAnnote}[2]{%
  \begin{quotation}\noindent\textsc{Key:} #1\\
  \textsc{Annotation:}\ #2\end{quotation}}

\bibitem[{Becker et~al.(2011)Becker, Minchenya, Zeidis, and Zimmermann}]{becker2011modeling}
Becker, F., Minchenya, V., Zeidis, I., and Zimmermann, K. (2011).
\newblock Modeling and dynamical simulation of vibration-driven robots.
\newblock In \emph{56th Internat. Scientific Colloquium Innovation in Mechanical Engineering: Shaping the Future}
\bibAnnoteFile{becker2011modeling}

\bibitem[{Calisti et~al.(2017)Calisti, Picardi, and Laschi}]{calisti2017fundamentals}
Calisti, M., Picardi, G., and Laschi, C. (2017).
\newblock Fundamentals of soft robot locomotion.
\newblock \emph{Journal of The Royal Society Interface} 14, 20170101
\bibAnnoteFile{calisti2017fundamentals}

\bibitem[{Cocuzza et~al.(2021)Cocuzza, Doria, and Reis}]{cocuzza2021vibration}
Cocuzza, S., Doria, A., and Reis, M. (2021).
\newblock Vibration-based locomotion of an amphibious robot.
\newblock \emph{Applied Sciences} 11, 2212
\bibAnnoteFile{cocuzza2021vibration}

\bibitem[{Golitsyna(2018)}]{golitsyna2018periodic}
Golitsyna, M. (2018).
\newblock Periodic regime of motion of a vibratory robot under a control constraint.
\newblock \emph{Mechanics of Solids} 53, 49--59
\bibAnnoteFile{golitsyna2018periodic}

\bibitem[{Gravish and Goldman(2014)}]{gravish2014effect}
Gravish, N. and Goldman, D.~I. (2014).
\newblock Effect of volume fraction on granular avalanche dynamics.
\newblock \emph{Physical Review E} 90, 032202
\bibAnnoteFile{gravish2014effect}

\bibitem[{Knuth et~al.(2012)Knuth, Johnson, Hopkins, Sullivan, and Moore}]{knuth2012discrete}
Knuth, M.~A., Johnson, J., Hopkins, M., Sullivan, R., and Moore, J. (2012).
\newblock Discrete element modeling of a mars exploration rover wheel in granular material.
\newblock \emph{Journal of Terramechanics} 49, 27--36
\bibAnnoteFile{knuth2012discrete}

\bibitem[{Lee et~al.(2022)Lee, Ryu, Kim, and Seo}]{lee2022compact}
Lee, K., Ryu, S., Kim, C., and Seo, T. (2022).
\newblock A compact and agile angled-spoke wheel-based mobile robot for uneven and granular terrains.
\newblock \emph{IEEE Robotics and Automation Letters} 7, 1620--1626
\bibAnnoteFile{lee2022compact}

\bibitem[{Li et~al.(2021)Li, Dutta, Cannon, Daymude, Avinery, Aydin et~al.}]{li2021programming}
Li, S., Dutta, B., Cannon, S., Daymude, J.~J., Avinery, R., Aydin, E., et~al. (2021).
\newblock Programming active cohesive granular matter with mechanically induced phase changes.
\newblock \emph{Science Advances} 7, eabe8494
\bibAnnoteFile{li2021programming}

\bibitem[{Liang et~al.(2012)Liang, Xu, Xu, Liu, Ren, Kong et~al.}]{liang2012amphihex}
Liang, X., Xu, M., Xu, L., Liu, P., Ren, X., Kong, Z., et~al. (2012).
\newblock The amphihex: A novel amphibious robot with transformable leg-flipper composite propulsion mechanism.
\newblock In \emph{2012 IEEE/RSJ International Conference on Intelligent Robots and Systems} (IEEE), 3667--3672
\bibAnnoteFile{liang2012amphihex}

\bibitem[{Maladen et~al.(2011)Maladen, Ding, Umbanhowar, and Goldman}]{maladen2011undulatory}
Maladen, R.~D., Ding, Y., Umbanhowar, P.~B., and Goldman, D.~I. (2011).
\newblock Undulatory swimming in sand: experimental and simulation studies of a robotic sandfish.
\newblock \emph{The International Journal of Robotics Research} 30, 793--805
\bibAnnoteFile{maladen2011undulatory}

\bibitem[{Marvi et~al.(2014)Marvi, Gong, Gravish, Astley, Travers, Hatton et~al.}]{marvi2014sidewinding}
Marvi, H., Gong, C., Gravish, N., Astley, H., Travers, M., Hatton, R.~L., et~al. (2014).
\newblock Sidewinding with minimal slip: Snake and robot ascent of sandy slopes.
\newblock \emph{Science} 346, 224--229
\bibAnnoteFile{marvi2014sidewinding}

\bibitem[{Notomista et~al.(2019)Notomista, Mayya, Mazumdar, Hutchinson, and Egerstedt}]{notomista2019study}
Notomista, G., Mayya, S., Mazumdar, A., Hutchinson, S., and Egerstedt, M. (2019).
\newblock A study of a class of vibration-driven robots: Modeling, analysis, control and design of the brushbot.
\newblock In \emph{2019 IEEE/RSJ International Conference on Intelligent Robots and Systems (IROS)} (IEEE), 5101--5106
\bibAnnoteFile{notomista2019study}

\bibitem[{Qian et~al.(2013)Qian, DAFFON, ZHANG, and Goldman}]{qian2013automated}
Qian, F., DAFFON, K., ZHANG, T., and Goldman, D.~I. (2013).
\newblock An automated system for systematic testing of locomotion on heterogeneous granular media.
\newblock In \emph{Nature-Inspired Mobile Robotics} (World Scientific). 547--554
\bibAnnoteFile{qian2013automated}

\bibitem[{Qian et~al.(2015)Qian, Zhang, Korff, Umbanhowar, Full, and Goldman}]{qian2015principles}
Qian, F., Zhang, T., Korff, W., Umbanhowar, P.~B., Full, R.~J., and Goldman, D.~I. (2015).
\newblock Principles of appendage design in robots and animals determining terradynamic performance on flowable ground.
\newblock \emph{Bioinspiration \& biomimetics} 10, 056014
\bibAnnoteFile{qian2015principles}

\bibitem[{Reis et~al.(2013)Reis, Yu, Maheshwari, and Iida}]{reis2013morphological}
Reis, M., Yu, X., Maheshwari, N., and Iida, F. (2013).
\newblock Morphological computation of multi-gaited robot locomotion based on free vibration.
\newblock \emph{Artificial life} 19, 97--114
\bibAnnoteFile{reis2013morphological}

\bibitem[{Rubenstein et~al.(2012)Rubenstein, Ahler, and Nagpal}]{rubenstein2012kilobot}
Rubenstein, M., Ahler, C., and Nagpal, R. (2012).
\newblock Kilobot: A low cost scalable robot system for collective behaviors.
\newblock In \emph{2012 IEEE international conference on robotics and automation} (IEEE), 3293--3298
\bibAnnoteFile{rubenstein2012kilobot}

\bibitem[{Shrivastava et~al.(2020)Shrivastava, Karsai, Aydin, Pettinger, Bluethmann, Ambrose et~al.}]{shrivastava2020material}
Shrivastava, S., Karsai, A., Aydin, Y.~O., Pettinger, R., Bluethmann, W., Ambrose, R.~O., et~al. (2020).
\newblock Material remodeling and unconventional gaits facilitate locomotion of a robophysical rover over granular terrain.
\newblock \emph{Science Robotics} 5, eaba3499
\bibAnnoteFile{shrivastava2020material}

\bibitem[{Tang et~al.(2018)Tang, Li, Fang, Li, and Chen}]{tang2018speedy}
Tang, C., Li, B., Fang, H., Li, Z., and Chen, H. (2018).
\newblock A speedy, amphibian, robotic cube: Resonance actuation by a dielectric elastomer.
\newblock \emph{Sensors and Actuators A: Physical} 270, 1--7
\bibAnnoteFile{tang2018speedy}

\bibitem[{Tegzes et~al.(2003)Tegzes, Vicsek, and Schiffer}]{tegzes2003development}
Tegzes, P., Vicsek, T., and Schiffer, P. (2003).
\newblock Development of correlations in the dynamics of wet granular avalanches.
\newblock \emph{Physical Review E} 67, 051303
\bibAnnoteFile{tegzes2003development}

\bibitem[{Wang et~al.(2022)Wang, Liu, Deng, Zhang, Li, Wang et~al.}]{wang2022miniature}
Wang, D., Liu, Y., Deng, J., Zhang, S., Li, J., Wang, W., et~al. (2022).
\newblock Miniature amphibious robot actuated by rigid-flexible hybrid vibration modules.
\newblock \emph{Advanced Science} , 2203054
\bibAnnoteFile{wang2022miniature}

\bibitem[{Zhong et~al.(2018)Zhong, Zhang, Xu, Zhou, Fang, and Li}]{zhong2018cpg}
Zhong, B., Zhang, S., Xu, M., Zhou, Y., Fang, T., and Li, W. (2018).
\newblock On a cpg-based hexapod robot: Amphihex-ii with variable stiffness legs.
\newblock \emph{IEEE/ASME Transactions on Mechatronics} 23, 542--551
\bibAnnoteFile{zhong2018cpg}

\end{thebibliography}





\end{document}